\documentclass[10pt,conference]{IEEEtran}

\usepackage{graphicx}
\usepackage{multirow}
\usepackage{listings}
\usepackage{cite}
\usepackage{booktabs}
\usepackage{tabularx}
\usepackage{amsmath,amssymb,amsfonts}
\usepackage{graphicx}
\usepackage{subcaption}
\usepackage{textcomp}
\usepackage{xcolor}
\usepackage{stfloats}
\usepackage{algorithm}
\usepackage{algpseudocode}
\usepackage[colorlinks=true,linkcolor=blue,citecolor=blue,urlcolor=blue]{hyperref}
\def\BibTeX{{\rm B\kern-.05em{\sc i\kern-.025em b}\kern-.08em
    T\kern-.1667em\lower.7ex\hbox{E}\kern-.125emX}}

\title{CAE: Character-Level Autoencoder for Non-Semantic Relational Data Grouping}
\author{
Veera V S Bhargav Nunna\textsuperscript{1}, 
Shinae Kang\textsuperscript{1}, 
Zheyuan Zhou\textsuperscript{1}, 
Virginia Wang\textsuperscript{2}, 
Sucharitha Boinapally\textsuperscript{3}, 
Michael Foley\textsuperscript{1} \\

\textsuperscript{1}Amazon Web Services Inc., Arlington, VA, USA \\
\textsuperscript{2}Amazon Web Services Inc., Seattle, WA, USA \\
\textsuperscript{3}Amazon Web Services Inc., Dallas, TX, USA \\

\{nunveera, kangshin, zheyuanz, vawang, bsuchi, folem\}@amazon.com
}

\begin{document}

\maketitle

% abstract
\begin{abstract} 

Enterprise relational databases increasingly contain vast amounts of non-semantic data—IP addresses, product identifiers, encoded keys, and timestamps—that challenge traditional semantic analysis. This paper introduces a novel Character-Level Autoencoder (CAE) approach that automatically identifies and groups semantically identical columns in non-semantic relational datasets by detecting column similarities based on data patterns and structures. Unlike conventional Natural Language Processing (NLP) models that struggle with limitations in semantic interpretability and out-of-vocabulary tokens, our approach operates at the character level with fixed dictionary constraints, enabling scalable processing of large-scale data lakes and warehouses. The CAE architecture encodes text representations of non-semantic relational table columns and extracts high-dimensional feature embeddings for data grouping. By maintaining a fixed dictionary size, our method significantly reduces both memory requirements and training time, enabling efficient processing of large-scale industrial data environments.
Experimental evaluation demonstrates substantial performance gains: our CAE approach achieved 80.95\% accuracy in top-5 column matching tasks across relational datasets, substantially outperforming traditional NLP approaches such as Bag of Words (47.62\%). These results demonstrate its effectiveness for identifying and clustering identical columns in relational datasets. This work bridges the gap between theoretical advances in character-level neural architectures and practical enterprise data management challenges, providing an automated solution for schema understanding and data profiling of non-semantic industrial datasets at scale.

\end{abstract}


% core paper
\section{Introduction}

In large enterprise environments, different teams often manage their own datasets independently, leading to a diverse and fragmented data landscape within enterprise data warehouses. As organizations scale, the number of relational tables grows rapidly, which in turn increases the number of the redundant or overlapping columns across different tables. These columns may represent the same or similar entities but their names or value formats are not standardized. This redundancy introduces significant challenges in identifying authoritative data sources, ensuring data consistency, and maintaining accurate data lineage.

To address these challenges, Natural Language Processing (NLP) techniques have been applied to automatically identify and group columns with semantic similarity across diverset set of relational tables. These approaches aim to capture the underlying semantics of column names and metadata, enabling tasks such as schema alignment, deduplication, and metadata standardization. Traditional models like Word2Vec \cite{word2vec} learn word-level embeddings by capturing co-occurrence patterns in text, producing dense vector representations that reflect semantic similarity. Another common method is Bag-of-Words (BoW)\cite{bagofwords}, which encodes text using one-hot vectors based on a predefined dictionary and applies the encoding result for comparison or classification. 

Despite their success in general NLP tasks, these models face two major limitations in enterprise data environments. First, the content of the column include non-semantic elements such as code names, IDs, or formatting artifacts do not convey meaningful information. Such non-semantic elements can mislead the model. Second, enterprise datasets frequently contain domain-specific terms, acronyms, and abbreviations that are not found in general-purpose vocabularies, resulting in out-of-vocabulary (OOV) \cite{outofvocabularychallengereport} issues and reduced embedding quality.

To overcome these limitations, we propose a Character-Level Autoencoder model tailored for relational column name embedding and similarity detection. By operating at the character level, our model avoids reliance on token boundaries, thus is more robust to noise, abbreviations, and rare or unseen terms. The autoencoder learns compact representations of column names by capturing character-level patterns that reflect column data similarity, even in the presence of spelling variations, domain-specific vocabulary, or formatting inconsistencies. In this paper, we describe the design of our model, the training methodology, and a set of experiments demonstrating its effectiveness in grouping redundant columns compared to existing baselines. Our results show that character-level modeling can provide meaningful improvements in identifying columns containing similar concepts.

\section{Related Work} 

\subsection{Dictionary-Based Language Model}
One of the predominant approaches in text embedding has been dictionary-based language models, which rely on predefined vocabularies to represent textual data. Traditional methods such as Bag-of-Words (BoW) \cite{bagofwords} represent documents as sparse vectors by encoding words through one-hot encoding against a fixed dictionary, where each dimension corresponds to the frequency of a specific word. More recently, transformer-based architectures like BERT \cite{bert} have revolutionized natural language processing by utilizing large-scale vocabularies (typically 30,000-50,000 tokens). BERT employs a multi-layer bidirectional transformer encoder that learns contextual representations through two primary tasks - masked language modeling and next sentence prediction - to achieve state-of-the-art performance on various language understanding tasks. However, all these sophisticated models present significant challenges when applied to non-semantic data: they require substantial computational resources, struggle with out-of-vocabulary tokens, and maintain large memory footprints due to their extensive vocabulary requirements.

\subsection{Semantic Language Model}
Semantic language models have revolutionized text representation by effectively capturing word relationships in continuous vector spaces. Word2Vec \cite{word2vec}, pioneered by Mikolov et al., established a fundamental approach to word embedding by training neural networks to learn word embeddings through two architectures: Continuous Bag-of-Words (CBOW), which predicts target words from context, and Skip-gram, which predicts context from target words. Building upon this foundation, FastText \cite{fasttext}, developed by Facebook Research, enhanced Word2Vec's methodology by incorporating subword information, thereby improving the handling of morphologically rich languages and out-of-vocabulary words. While these semantic models have significantly advanced natural language processing tasks, they prove inadequate for processing non-semantic data that is increasingly prevalent in industrial databases, such as network IP addresses, product identifiers, and other structured identifiers where semantic relationships are not meaningful or applicable.

\subsection{Character-Level Text Processing and Recent Advances}
Character-level text processing offers a powerful alternative to word-based approaches, particularly for structured data where traditional tokenization fails. Zhang et al. \cite{ZhangZhaoLeCun2015} demonstrated that character-level convolutional networks achieve competitive performance by treating text as a one-dimensional signal without relying on word-level semantics. This foundation has evolved through architectural innovations: Johnson and Zhang \cite{johnson2017deep} introduced Deep Pyramid CNNs that improved computational efficiency, while Huang and Wang \cite{huang2016character} successfully adapted character-level CNNs to non-alphabet languages. Zampieri et al. \cite{zampieri2017word} further demonstrated the robustness of character-level models for noisy text, showing superior language-agnostic performance compared to word-based approaches.

Hybrid architectures have extended these capabilities through innovative designs. Lai et al. \cite{lai2015recurrent} proposed Recurrent Convolutional Neural Networks that combine character and word-level information. Al-Rfou et al. \cite{al2020evolving} employed evolutionary algorithms to automatically discover optimal character-level CNN architectures—approaches particularly valuable for processing non-semantic data common in enterprise databases.

Recent advances have further refined character-level modeling for specialized applications. Van den Bosch et al. \cite{vandenbosch2023what} established benchmarks for character-level encoder architectures optimal for specialized identifiers, while Li et al. \cite{li2024empowering} addressed sub-token conflicts through their Fill-In-the-Middle approach. Xu et al. \cite{xu2024enhancing} enhanced character-level understanding in language models by incorporating token internal information. The versatility of these approaches extends to diverse applications, including Cao et al.'s \cite{cao2024chargen} enhancement of visual text generation.

Most relevant to our work is the integration of character-level processing with autoencoder frameworks, exemplified by Cunningham et al. \cite{cunningham2024sparse}, who showed how autoencoders can decompose complex representations into interpretable features. Our Character-Level Autoencoder builds on this research trajectory, applying character-level autoencoding specifically to non-semantic relational data to address critical gaps in enterprise data management systems processing structured information.

\subsection{Auto-Encoder Architectures for Text Compression}
Autoencoders \cite{autoencoder} represent a foundational class of unsupervised learning models for dimensionality reduction and feature learning. The architecture comprises two key components: an encoder that compresses input data into dense latent space; and a decoder that reconstructs the original data from this compressed representation. This approach has proven particularly effective for learning compact, meaningful representations while filtering out noise. Among various autoencoder variants, convolutional autoencoders \cite{marzougui2020data, al-ajmi2021binary} specifically excel at processing structured input data by employing convolutional layers, which preserve spatial relationships during the encoding-decoding process. The proposed models using an embedding layer and a convolutional layer offer high accuracy in various classification tasks.

The convergence of character-level processing with autoencoder architectures presents particular advantages for non-semantic data processing. The convolutional networks were proven effective for classifying texts by treating language as a signal, rather than relying on semantics \cite{ZhangZhaoLeCun2015}. Drawing from Prusa and Khoshgoftaar's character-level encoding framework \cite{cle} and recent developments in textual data classification \cite{lakshmi2022efficient}, autoencoder-based approaches can effectively compress character-level representations while maintaining the structural patterns essential for data matching and grouping tasks. This combination proves especially valuable for industrial database applications where semantic relationships are less important than structural similarity patterns.

Our work adopts this architecture to process character-level text embeddings, leveraging its ability to capture local patterns and structural features for generating condensed representations for efficient data grouping. Our approach first transforms each character in table columns using 1-of-$m$ encoding (where $m$ represents the character alphabet size), creating a binary vector representation that preserves character-level patterns. These representations are then processed through an autoencoder architecture that compresses the character-level features into compact, fixed-dimension column embeddings. This enables efficient data grouping while maintaining the structural integrity of non-semantic data, demonstrating how theoretical advances from Zhang et al. can be scaled to address practical challenges in enterprise data environments.
\section{Character-Level Auto-Encoder}

\begin{figure}
    \centering
    \includegraphics[width=1\linewidth]{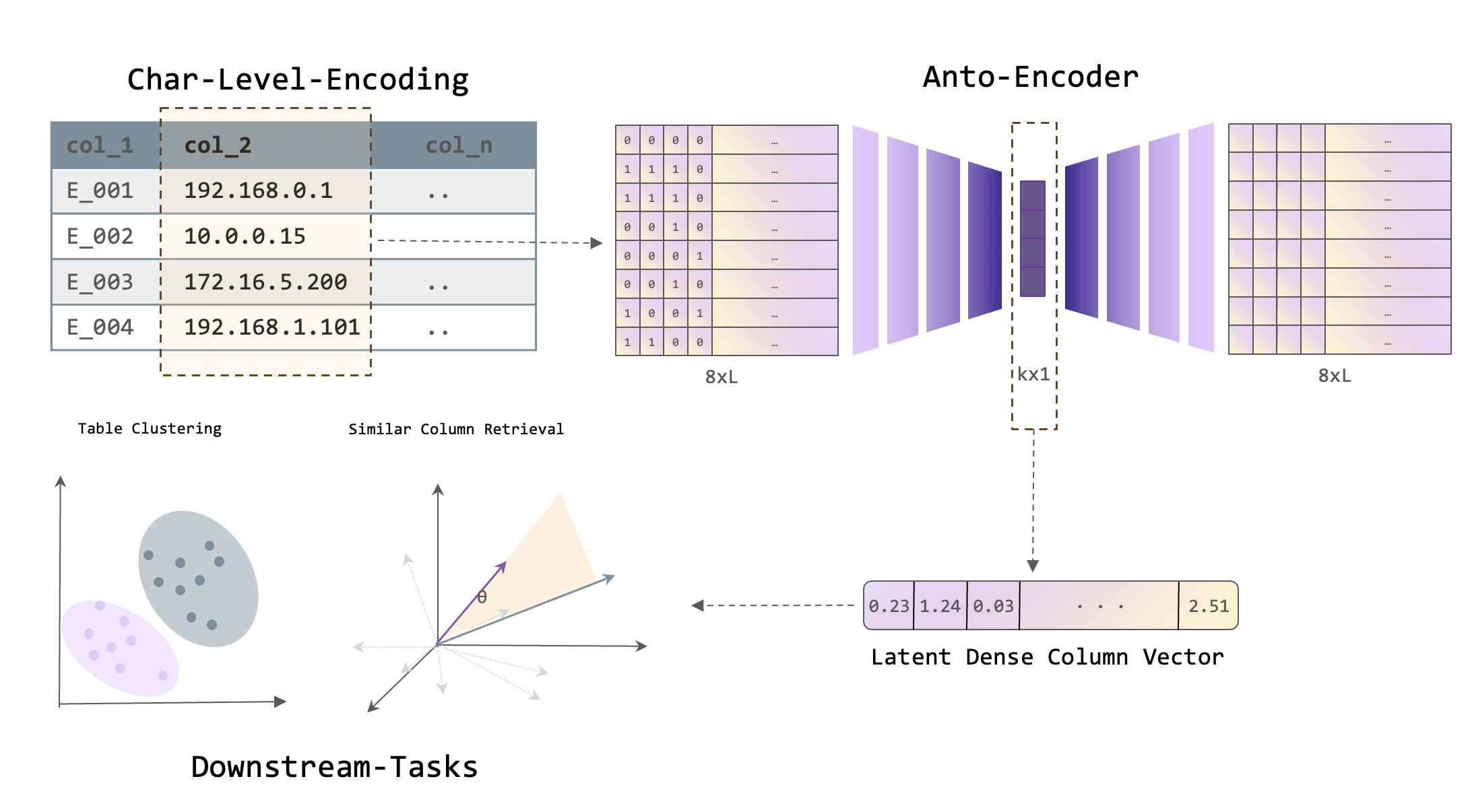}
    \caption{Character-Level Auto-Encoder framework: (1) Character-Level Encoding converts table column text into sparse matrices; (2) Auto-Encoder compresses these matrices into dense latent vectors and reconstructs the original encoding; (3) Latent vectors enable column grouping via cosine similarity measurement.}
    \label{fig:full_experiment}
\end{figure}

The proposed Character-Level Auto Encoder (CAE) is split into two parts: Character-Level-Encoding (CLE) and Auto-Encoder. The first component, Character-Level Encoding, processes relational table column data by transforming each character into a one-hot encoded vector, where each position in the vector corresponds to a unique character in the alphabet. The second component, Auto-Encoder, takes the sparse character-level encoded matrix as input and performs two sequential operations: [1] The encoder compresses the sparse representation into a dense hidden feature vector; [2] The decoder attempts to reconstruct the original character-level encoding from this compressed representation. Through this two-stage process, our model learns to capture essential patterns in non-semantic data while significantly reducing the dimensionality of the representation (Fig. \ref{fig:full_experiment}).
    
\subsection{Character-Level-Encoding}
\label{sebsec:CLE}
Zhang and LeCun proposed a 1-to-$m$ character-level encoding, where each character is represented by a binary vector of length $m$ ($m$ being the alphabet size), with a value of 1 at the position corresponding to the character and 0 elsewhere\cite{ZhangZhaoLeCun2015}. Our CAE model adopted this method into relational table data by choosing ASCII code \cite{ascii} as our character set. This algorithm takes a table and one of its columns as input. For each entry in the column, it encodes characters up to a fixed cutoff length into their 8-bit ASCII binary vectors. If an entry has fewer characters than the cutoff, it pads the encoding with zero vectors to maintain a consistent length. Each encoded entry becomes a fixed-size row vector, and all such vectors stack together to form a matrix representing the entire column. This process can be described as the following pesudo code:

\begin{algorithm}
\caption{Character-Level Encoding of Table Column}
\begin{algorithmic}[1]
\Require Table $T$, column $C$, cutoff length $L$
\State Initialize matrix $M \gets []$
\ForAll{entry $e$ in column $C$ of table $T$}
    \State Initialize encoded list $E \gets []$
    \For{$i = 1$ to $\min(\text{length}(e), L)$}
        \State $c \gets$ character at position $i$ in $e$
        \State $a \gets$ 8-bit ASCII encoding of $c$
        \State Append $a$ to $E$
    \EndFor
    \While{$|E| < L$}
        \State Append zero vector $\mathbf{0}_{1 \times 8}$ to $E$
    \EndWhile
    \State Append $E$ as a new row to matrix $M$
\EndFor
\State \Return matrix $M$
\end{algorithmic}
\end{algorithm}

% talk about it can be generalized to UTF-8 encoding
While our implementation utilizes ASCII encoding due to the English-based nature of our dataset, this Character-Level-Encoding framework can be readily extended to support other character encoding systems. For instance, the method can be adapted to use UTF-8 encoding by simply modifying the character-to-binary conversion step, allowing the system to handle multilingual content and special characters while maintaining the same overall encoding structure. This flexibility makes the framework suitable for diverse datasets requiring different character encoding schemes.

\begin{figure}[htbp]
  \centering
  \begin{subfigure}{0.5\textwidth}
    \includegraphics[width=\linewidth]{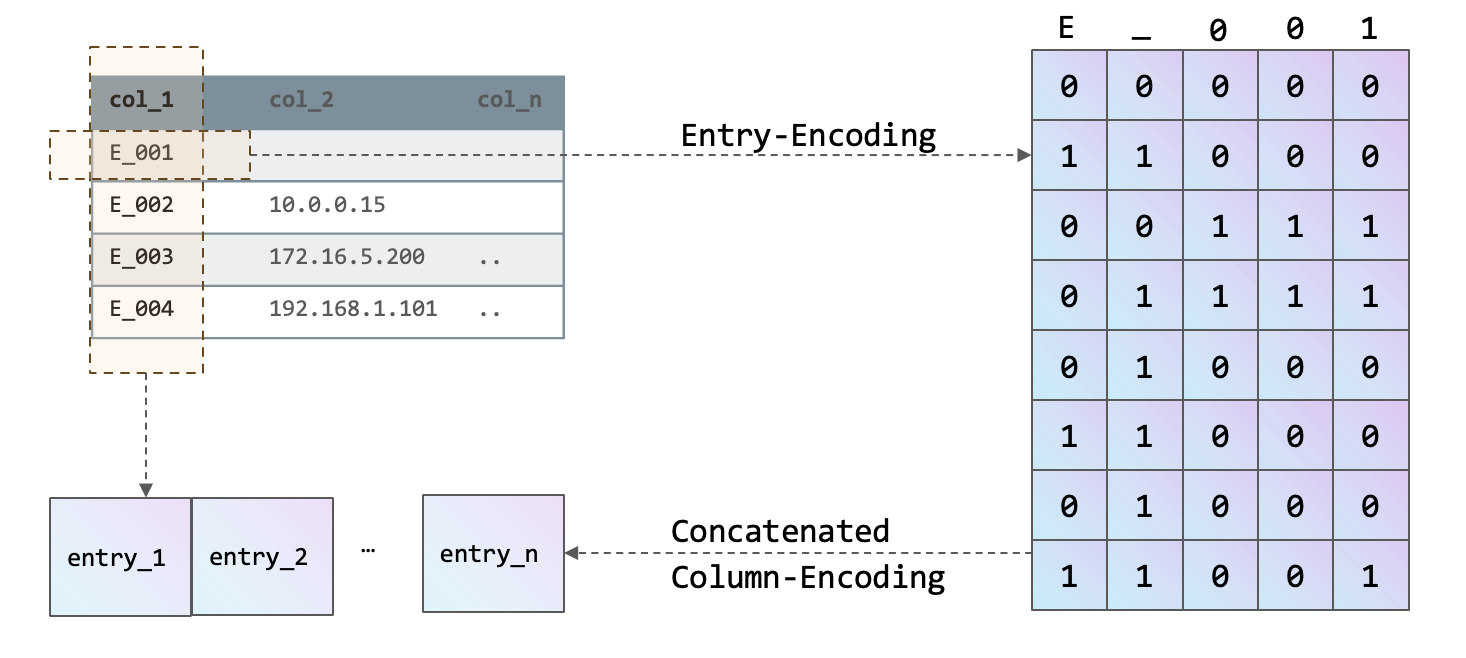}
    \caption{Concatenated CLE}
    \label{fig:img1}
  \end{subfigure}
  \hfill
  \begin{subfigure}{0.5\textwidth}
    \includegraphics[width=\linewidth]{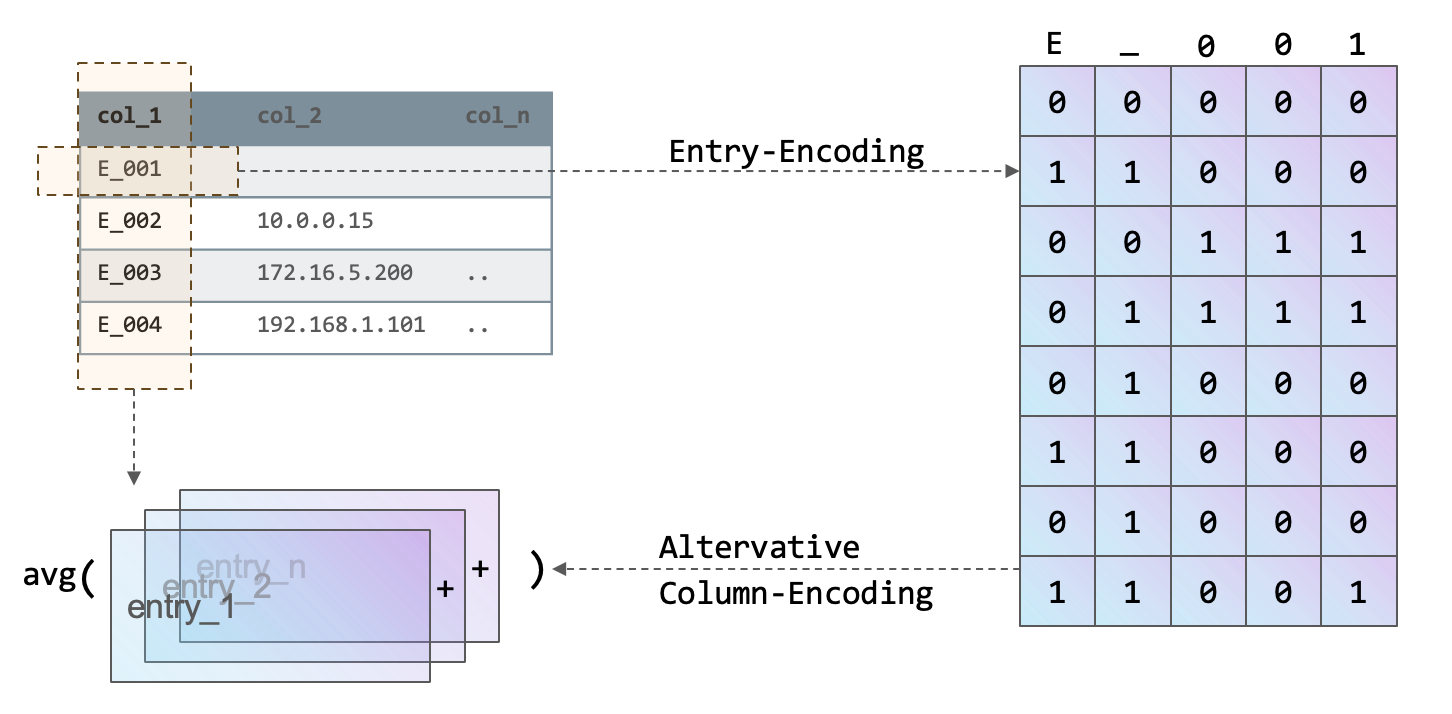}
    \caption{Alternative CLE}
    \label{fig:img2}
  \end{subfigure}
  \caption{Two character-level encoding (CLE) approaches for column vector assembly: (a) Concatenated encoding sequentially joins entry vectors up to a length limit, and (b) Alternative CLE averages entry vectors to create a smoothed representation.}
  \label{fig:twocle}
\end{figure}

We evaluated two methods to aggregate entry vectors into a column vector of shape (8,L), where L is a predefined cutoff (Fig. \ref{fig:twocle}). The Concatenated-CLE method concatenates entry encodings sequentially until reaching the length limit or the end of the column. The Alternative-CLE method computes the average of all entry encodings. Based on experimental results, we recommend Alternative-CLE, as it captures more comprehensive content information within the same output shape. Additionally, averaging the entry encodings helps smooth the feature vector by reducing noise.

\subsection{Auto-Encoder}
After retrieving sparse column encoding matrix from character-level encoding, we feed the matrix into the autoencoder to obtain a dense feature vector and use it for feature data autogrouping. Let \( M \in \mathbb{R}^{8 \times L} \) be the input matrix obtained from character-level encoding. Then, the general autoencoder model can be noted as follows:

\textbf{Encoder:} The encoder maps the input matrix \( M \) into a latent vector \( Z \in \mathbb{R}^k \), where \( W_e \in \mathbb{R}^{k \times (8L)} \), \( b_e \in \mathbb{R}^{k} \), and \( \text{vec}(M) \in \mathbb{R}^{8L} \) denotes flattening \( M \) into a vector. \( \phi \) is a non-linear activation function:
\begin{equation}
    Z = f_{\text{enc}}(M) = \phi(W_e \cdot \text{vec}(M) + b_e)
\end{equation}

\textbf{Decoder:} The decoder reconstructs the input matrix from the latent vector, where \( W_d \in \mathbb{R}^{(8L) \times k} \), \( b_d \in \mathbb{R}^{8L} \), and \( \hat{M} \in \mathbb{R}^{8 \times L} \) is reshaped from the output:
\begin{equation}
    \hat{M} = f_{\text{dec}}(Z) = \psi(W_d Z + b_d)
\end{equation}

\textbf{Loss:} We minimize the reconstruction error using Mean Squared Error (MSE), where \( \| \cdot \|_F \) denotes the Frobenius norm:
\begin{equation}
    \mathcal{L}_{\text{AE}} = \| M - \hat{M} \|_F^2
\end{equation}

We implemented and evaluated two types of autoencoders: a Linear Autoencoder (Linear AE) and a Convolutional Autoencoder (Convolutional AE). The Linear AE consists of an encoder and decoder built entirely from fully connected layers. The encoder compresses the input through a series of linear layers, gradually reducing its dimensionality to a 100-dimensional latent vector, with dropout at a rate of 0.2 applied for regularization. The decoder reconstructs the original input by symmetrically expanding the latent vector through linear layers back to the input dimension.

In the Convolutional AE, the encoder starts with two convolutional layers that increase and maintain 16 channels using \(3 \times 3\) kernels with same padding. The resulting feature maps are flattened and passed through fully connected layers with ReLU activations, compressing the data into a 100-dimensional latent vector. The decoder reverses this process by expanding the latent vector via fully connected layers, reshaping it into feature maps, and applying two convolutional layers that reduce the channels back to 1, reconstructing the original input.

\section{Experiment and Result}

\begin{figure}[htbp]
  \centering
  \begin{subfigure}{0.45\textwidth}
    \includegraphics[width=\linewidth]{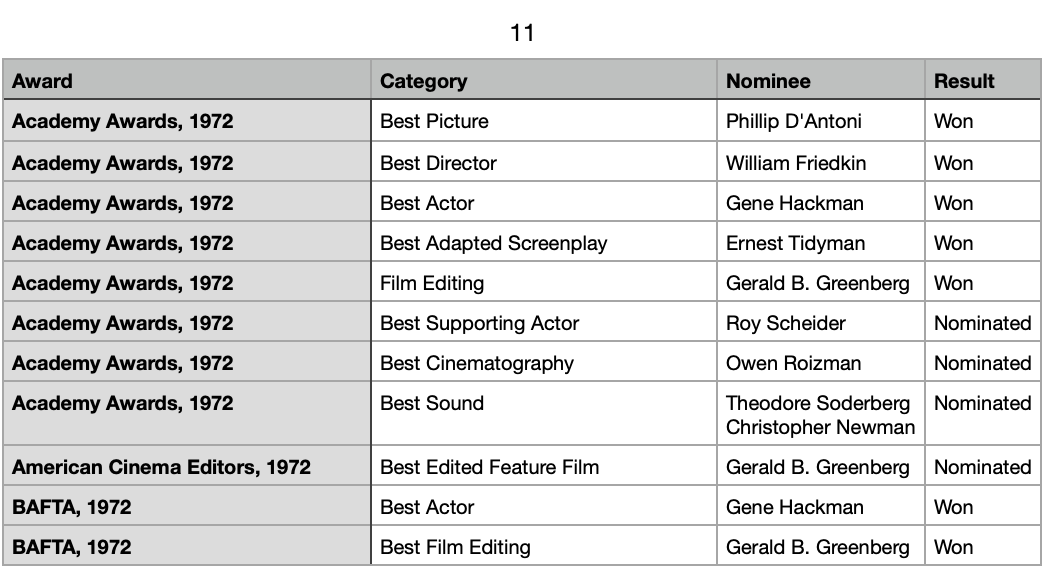}
    \caption{WikiTableQuestions Table 11}
    \label{fig:img1}
  \end{subfigure}
  \hfill
  \begin{subfigure}{0.45\textwidth}
    \includegraphics[width=\linewidth]{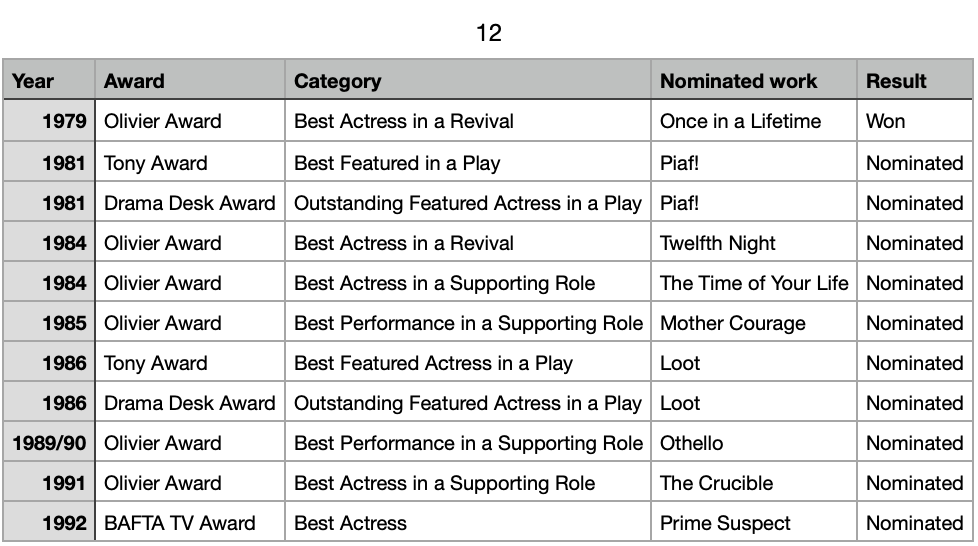}
    \caption{WikiTableQuestions Table 12}
    \label{fig:img2}
  \end{subfigure}
  \caption{Sample WikiTableQuestions tables that should be grouped together on shared  Award, Category, or Result columns.}
  \label{fig:exmaplegrouptable}
\end{figure}

\subsection{Dataset} 

For this experiment, we use the WikiTableQuestions dataset \cite{pasupat2015compositionalsemanticparsingsemistructured}, which contains 2,108 semi-structured HTML tables extracted from Wikipedia. TThe dataset comprises alphanumeric characters, making it relevant for testing the extent of character-level encoding. On average, each table has 6.3 columns and 27.5 rows. The dataset includes a wide variety of table structures and content, making it suitable for evaluating column similarity methods. A simple example is shown in Fig. \ref{fig:exmaplegrouptable}: Table 11 focuses on an actress’ filmography and table 12 focuses on a specific year's film awards. Both contain columns related to film awards. Columns such as Award, Category, or Result may be clustered together by the model, despite the structural and contextual differences between the tables. 

\subsection{Data Preprocessing}
\begin{figure}
    \centering
    \includegraphics[width=1\linewidth]{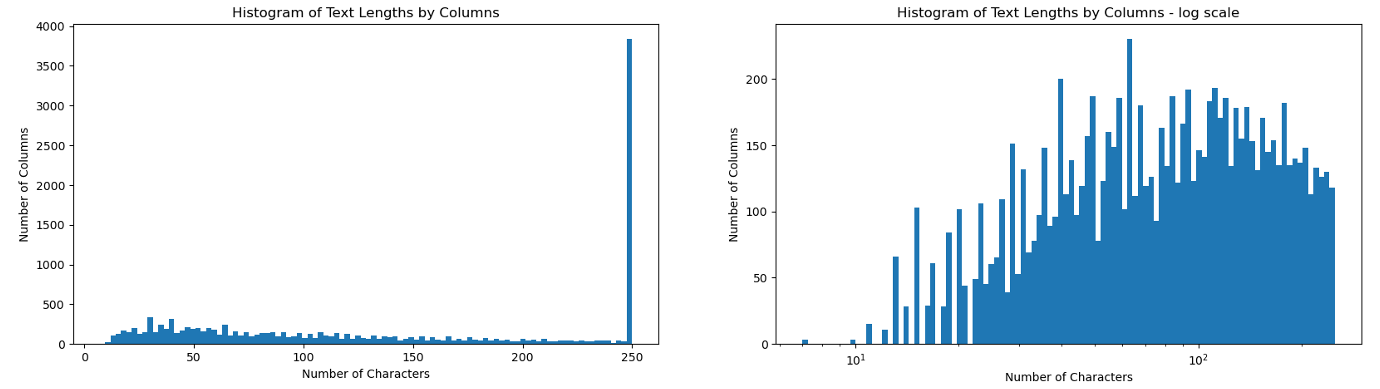}
    \caption{Text length distributions across dataset columns: linear scale (left) and logarithmic scale (right), with the spike at 250 corresponding to the selected character cutoff threshold.}
    \label{fig:table_statistic}
\end{figure}

We partitioned the dataset into training, validation and testing sets using an 70/20/10 split across all models.
During preprocessing, we addressed non-English characters not supported by extended ASCII encoding by replacing unrecognizable characters with null characters (zero vectors). 
To maintain consistent input dimensions, we set a fixed text length, $TEXT\_CUTOFF$, for each input. The cutoff length was determined based on dataset statistics (Table. \ref{tab:text_stats}) to capture the majority of columns while keeping the length reasonable. 
% -------table start
\begin{table}[h]
\centering
\begin{tabular}{lc}
\toprule
\textbf{Statistic} & \textbf{Value} \\
\midrule
Mean & 314.38 \\
Standard deviation & 822.16 \\
Variance & 675,942.16 \\
Minimum & 7 \\
Maximum & 24,968 \\
\bottomrule
\end{tabular}
\caption{Text Length Statistics}
\label{tab:text_stats}
\end{table}
Using a dynamic threshold sliding window to calculate the column count, we found that 71.68\% of the columns were fully preserved without truncation with $TEXT\_CUTOFF$ = 250  (Fig. \ref{fig:table_statistic}).
% -------figure start

\subsection{Setup}
For our experiments, we established four distinct models that combine different character encoding schemes and autoencoder architectures, as summarized in Table \ref{tab:models}. To ensure consistency and comparability across all models, we standardized several key training parameters: the dataset was partitioned with 80\% allocated for training and 20\% reserved for validation; the learning rate was uniformly set to 0.001; each model was trained for 100 epochs with a batch size of 64. Additionally, mean squared error (MSE) loss was employed as the criterion for optimization, and the Adam optimizer \cite{kingma2015adam} was utilized to update the network weights effectively.
\begin{table}[h]
\centering
\begin{tabular}{l c l}
\toprule
\textbf{Model Notation} & \textbf{Character Encoding} & \textbf{Auto-encoder}       \\
\midrule
Concatenated Linear       & Concatenated             & Linear           \\
Concatenated Convolution  & Concatenated             & Convolution    \\
Alternative Linear        & Alternative & Linear           \\
Alternative Convolution & Alternative & Convolution    \\
\bottomrule
\end{tabular}
\caption{Comparison of Four Model Variants}
\label{tab:models}
\end{table}

\subsection{Evaluation}
\label{subsec:evaluation}
\subsubsection{Reconstruction}
\begin{figure}[h]
\centering
\begin{subfigure}{\columnwidth}
    \centering
    \includegraphics[width=0.9\columnwidth]{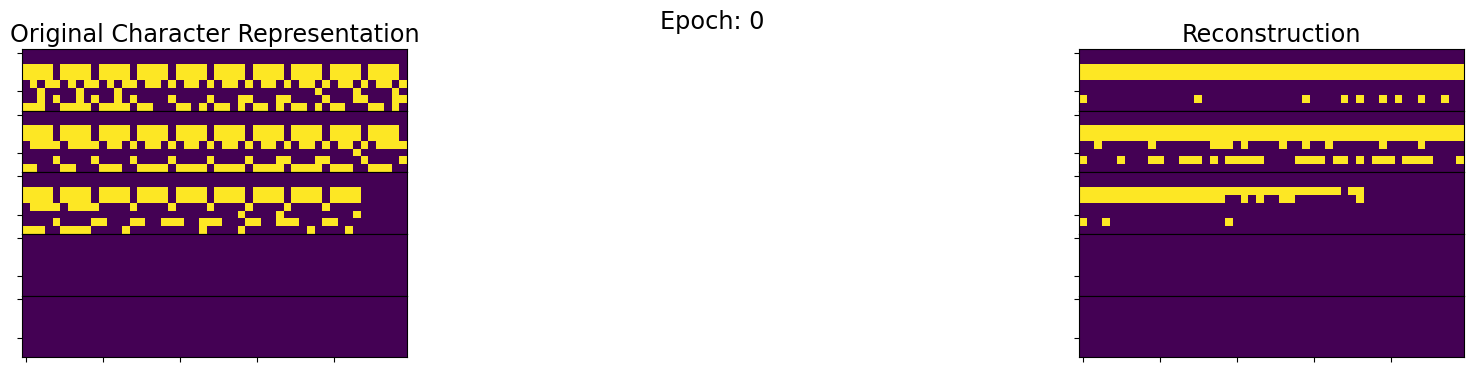}
\end{subfigure}

\begin{subfigure}{\columnwidth}
    \centering
    \includegraphics[width=0.9\columnwidth]{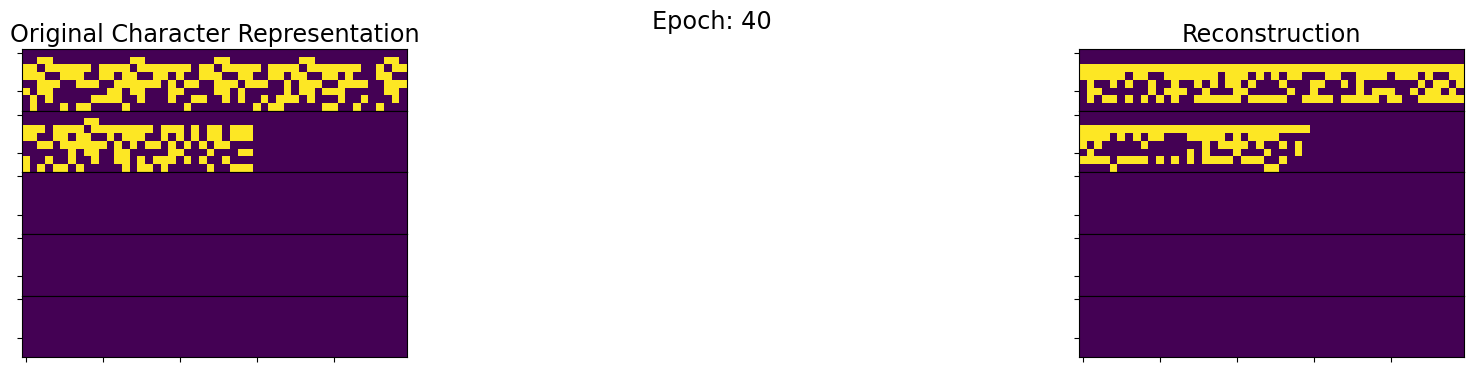}
\end{subfigure}

\begin{subfigure}{\columnwidth}
    \centering
    \includegraphics[width=0.9\columnwidth]{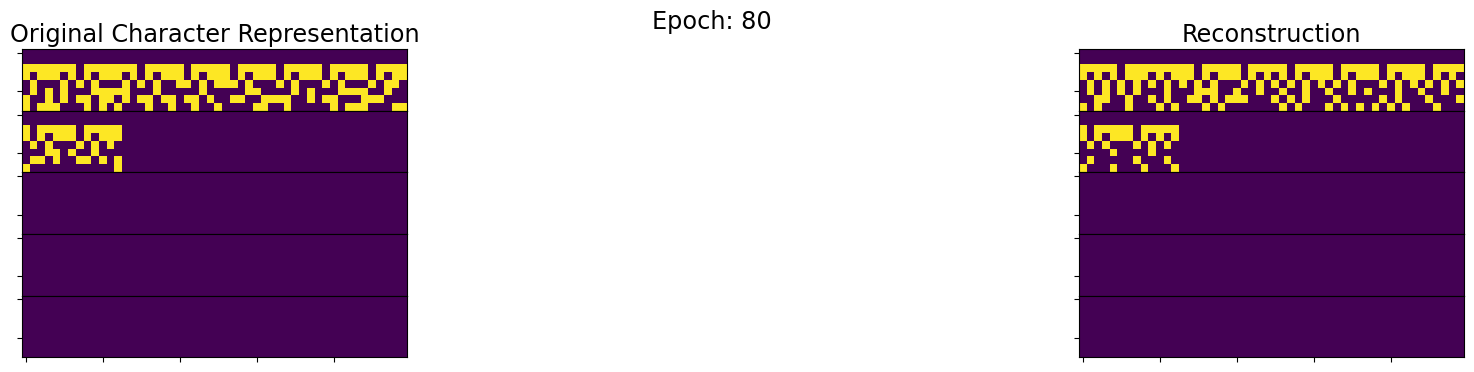}
\end{subfigure}

\caption{Reconstruction quality improvement over training epochs: Input encoding matrix (left) and their reconstructions (right). As the number of training epochs increases, the model progressively captures salient features and underlying patterns, resulting in reconstructions that more closely resemble the original encoded inputs.}
\label{fig:reconstruction_vis}
\end{figure}

With the above experimental setup, we monitored the training process by visualizing both the input character-encoded vectors and their corresponding reconstructed vectors, as illustrated in Fig.~\ref{fig:reconstruction_vis}. As the number of training epochs increased, the reconstructions progressively aligned more closely with the original inputs, indicating that the model was successfully learning to capture and reproduce the underlying features of the encoded representations. This trend reflects the model’s gradual convergence, demonstrating improved reconstruction accuracy over time.

\subsubsection{Clustering}
\begin{figure}
    \centering
    \includegraphics[width=0.8\columnwidth]{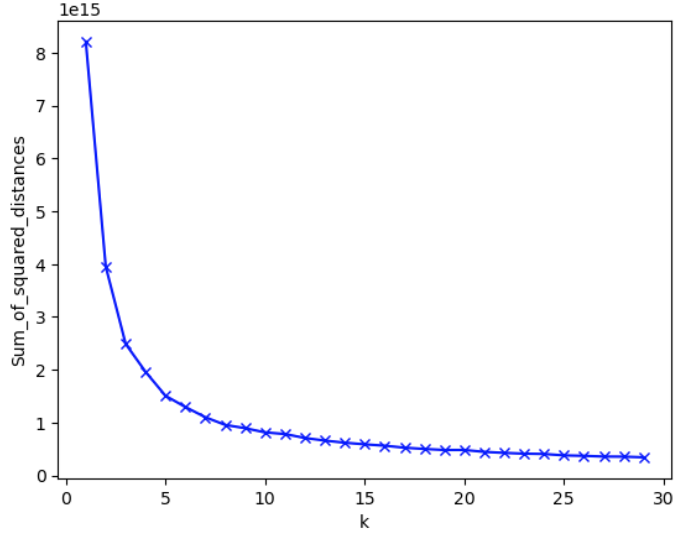}
    \caption{Elbow method plot for determining the optimal number of clusters showing 6 being the optimal cluster.}
    \label{fig:elbow_method}
\end{figure}

\begin{figure}
    \centering
    \includegraphics[width=0.8\columnwidth]{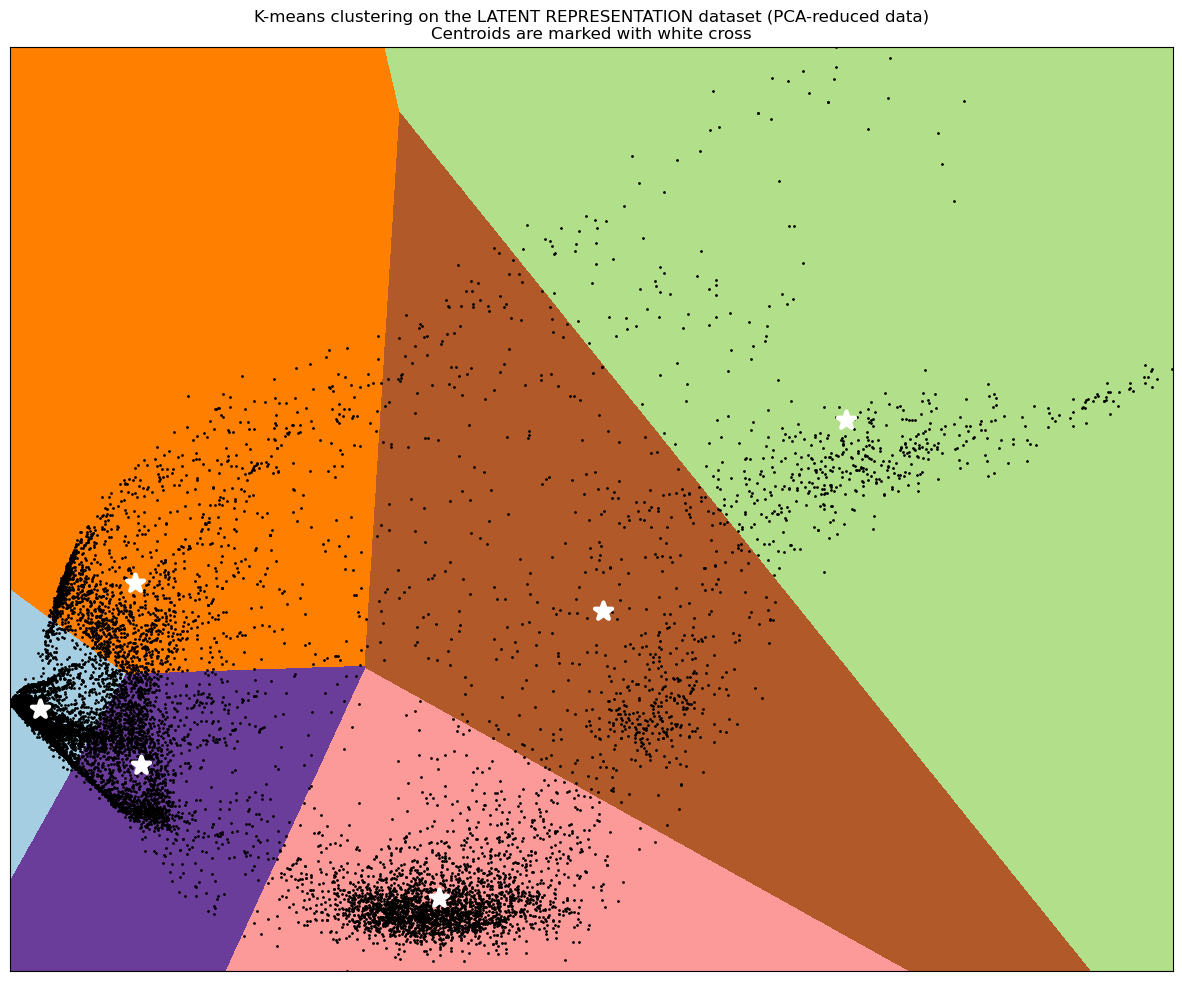}
    \caption{K-means clustering on CAE output WikiTableQuestions column latent representation with PCA reduction.}
    \label{fig:clustering}
\end{figure}

After completing model training, we applied k-means clustering \cite{macqueen1967kmeans} to the column embeddings to obtain a high-level view of the relationships among their latent vector representations. Due to the diverse nature of the WikiTableQuestions dataset \cite{pasupat2015compositionalsemanticparsingsemistructured}, determining the optimal number of clusters presented a challenge. To address this, we employed the elbow method \cite{james2013islr}, which evaluates a range of candidate cluster counts and computes the within-cluster sum of squares (WCSS) for each. The optimal number of cluster was determined at the “elbow” point where adding additional clusters yields diminishing improvements in WCSS. Our experiments revealed this point at six clusters, as shown in Fig.~\ref{fig:elbow_method}.
Using the determined six clusters, we applied Principal Component Analysis (PCA) \cite{pearson1901liii} for dimensionality reduction to project the high-dimensional data into a lower-dimensional space. As shown in Fig.~\ref{fig:clustering}, the resulting visualization reveals distinct groupings that correspond to the identified clusters. The clear separation between groups suggests that the model successfully captured the inherent features of the data.

\subsection{Result}
For our evaluation, we reserved 10\% of the complete dataset as testing data. Within this test set, we manually curated a collection of column pairs from our database tables, focusing specifically on non-semantic data types such as identifiers, codes, and reference numbers. For example, we included pairs like "Coordinates" (index: 201-csv/14, 201-csv/2). This testing strategy allowed us to evaluate our model's ability to effectively group non-semantic data types while distinguishing them from semantic content like names or descriptions. Building on the qualitative findings presented in Section~\ref{subsec:evaluation}, we conducted a quantitative evaluation of our model by measuring top-k hits \cite{fagin2003optimal} using cosine similarity \cite{salton1988term} across four variations of our convolutional autoencoder (CAE) models and baseline implementations. 

The comparative results indicate that the alternative convolutional autoencoder consistently outperforms the other models. In particular, the alternative encoding method generally achieves superior retrieval performance compared to the concatenated encoding. This improvement is likely attributed to the aggregation operation employed in the alternative approach. This aggregation smooths potential noise across entries in all rows and enhances the model’s ability to capture underlying column patterns. Furthermore, convolutional autoencoders demonstrate better performance than linear autoencoders, which may be attributed to the convolutional kernels’ capability to capture spatial information within the constructed column encoding matrix, such as the local neighborhood of characters that provide richer feature representations.
\begin{table}[h]
\caption{Top-k Performance Comparison for Different CAE models}
\label{tab:topk_performance}
\centering
\setlength{\tabcolsep}{10pt} % adjust column separation if needed
\renewcommand{\arraystretch}{1.5} % adjust row height for better spacing
\begin{tabular}{lcccc}
\hline
\multirow{2}{*}{\textbf{CLE Type}} & \multicolumn{2}{c}{\textbf{Linear AE}} & \multicolumn{2}{c}{\textbf{Convolution AE}} \\
\cline{2-5}
 & Top 1 & Top 5 & Top 1 & Top 5 \\
\hline
Concatenated & 52.38\% & 71.43\% & 73.91\%  & 78.26\% \\
Alternative  & 71.43\% & 80.95\% & 76.19\%  & \textbf{85.71\%} \\
\hline
\end{tabular}
\end{table}

We evaluated our approach using an alternative convolutional autoencoder as our representative CAE model, comparing its performance against two baseline models: Bag-of-Words and Word2vec. As shown in Table~\ref{tab:topk_baseline_comparison}, the CAE significantly outperforms both baselines, achieving a Top-1 accuracy of 76.19\% and a Top-5 accuracy of 85.71\%, respectively. The baseline methods demonstrate substantially lower performance: BoW achieved Top-1 accuracies of 4.76\% and Top-5 accuracies of 47.62\%, while Word2vec reached and 19.05\% and 52.38\%, respectively. These results underscore the CAE's superior capability in capturing meaningful and discriminative features for retrieval tasks, attributed to its ability to learn hierarchical and spatial representations that traditional encoding methods cannot capture.

\begin{table}[h]
\caption{Top-k Performance Comparison of CAE and Baseline Models}
\label{tab:topk_baseline_comparison}
\centering
\setlength{\tabcolsep}{10pt}
\renewcommand{\arraystretch}{1.5}
\begin{tabular}{lcc}
\hline
\textbf{Encoding Method} & \textbf{Top 1} & \textbf{Top 5} \\
\hline
CAE               & 76.19\% & \textbf{85.71\%} \\
BoW               & 4.76\%  & 47.62\% \\
Word2vec          & 19.05\% & 52.38\% \\
\hline
\end{tabular}
\end{table}

\section{Discussion}

Our experimental results demonstrate that the Character-Level Autoencoder (CAE) approach significantly outperforms traditional semantic text embedding techniques for identifying similar columns in relational datasets. The substantial performance gap — with our best CAE model achieving 85.71\% Top-5 accuracy compared to Word2vec's 52.38\% and BoW's 47.62\%—indicates the fundamental limitations of conventional NLP approaches when processing structured tabular data with varied column formats.

\subsection{Analysis of Encoding and Architecture Choices}

The comparative results revealed that the alternative encoding method consistently achieves superior retrieval performance compared to the concatenated encoding approach (85.71\% vs. 80.95\% Top-5 accuracy with linear autoencoders, and 85.71\% vs. 78.26\% with convolutional autoencoders). This improvement can be attributed to the aggregation operation employed in the alternative approach, which effectively smooths potential noise across entries in all rows and enhances the model's ability to capture underlying column patterns despite variations in individual values.

Furthermore, convolutional autoencoders demonstrated better performance than linear autoencoders across both encoding methods. This superiority likely stems from the convolutional kernels' capability to capture spatial information within the constructed column encoding matrix. The 2D convolutions can identify local neighborhoods of characters that provide richer feature representations, effectively learning patterns that exist across structured identifiers and codes found in the diverse WikiTableQuestions dataset.

\subsection{Clustering Effectiveness}

The clustering results visualized in Fig.~\ref{fig:clustering} reveal distinct groupings that correspond to natural categories within the WikiTableQuestions dataset. The elbow method's identification of six optimal clusters, as shown in Fig.~\ref{fig:elbow_method}, suggests that columns naturally organize into this number of broad categories based on their character-level patterns. The clear separation between clusters indicates that the model successfully captured inherent features of the data even without explicit semantic understanding.

Upon examination of cluster contents, we observed that columns with similar data types and functions (such as date fields, identifiers, and categorical labels) were consistently grouped together despite variations in their representation across different tables. This emergent organization demonstrates the effectiveness of character-level pattern recognition for column similarity detection in heterogeneous data environments like the WikiTableQuestions dataset, which spans diverse topics and table structures.

\section{Caveat}

Although our Character-Level Autoencoder approach demonstrates significant improvements over traditional methods, we acknowledge several limitations in our current study. First of all, our comparative analysis focused primarily on basic baselines (BoW and Word2Vec) rather than more recent advanced embedding techniques like BERT or domain-specific language models. This choice was deliberate, as our research aimed to establish the fundamental advantages of character-level processing for structured data rather than competing with state-of-the-art semantic models. Additionally, our experiments were tested on a single dataset (WikiTableQuestions). While the size may be adaquate, the range of topics covered might not fully represent the diversity found in actual enterprise environments. Finally, though our model shows strong performance, the current implementation offers limited interpretability into how specific character patterns influence column similarity judgments.
\section{Conclusion and Future Work}

This paper presented a Character-Level Autoencoder (CAE) approach to address the critical challenge in handling complex naming variations and semantic ambiguities, a problem prevalent in enterprise data warehouses. Our proposed method overcomes limitations of traditional NLP techniques by operating at the character level, avoiding dependency on token boundaries and effectively handling non-semantic elements, domain-specific term, and out-of-vocabulary issues. %Let's delete this if this feels redundant 
Specifically, our approach employs ASCII-based character encoding (readily extendable to UTF-8 support) combined with an averaging mechanism that aggregates character-level representations across column entries, enabling the autoencoder to learn robust, noise-resistant embeddings that capture semantic similarity even in the presence of formatting variations and domain-specific terminology. 

Through experiments on the WikiTableQuestions Dataset, we demonstrate that Alternative convolutional Autoencoder achieves superior performance with 76.19\% Top-1 and 85.71 \% Top-5 accuracy, significantly outperforming baseline methods including BoW (4.76\% Top-1) and Word2Vec (19.05 \% Top-1).

This work not only substantially improves schema matching in traditional data warehouse environments but also addresses fundamental data governance challenges in large-scale enterprise systems. As organizations continue to accumulate vast amounts of data across distributed teams and systems, the ability to automatically identify semantic relationships between columns becomes crucial for maintaining data quality, ensuring consistency, and establishing accurate data lineage. Our character-level approach proves particularly valuable in handling ambiguity and complexity inherent in real-world enterprise datasets, where standardized naming conventions are often absent and domain-specific acronyms and abbreviations are common. 

In Big Data contexts, our approach offers a practical solution for metadata management and data discovery at scale. This model's robustness to spelling variations, formatting inconsistencies, and rare terms make it well-suited for management and upkeep of massive and heterogeneous datasets. Future work will focus on three key directions: (1) evaluating the model's scalability across larger enterprise datasets; (2) exploring semi-supervised or contrastive learning frameworks to further improve feature discriminability between similar columns; and (3) enhancing model interpretability through visualization of reconstruction errors and learned embedding spaces, providing greater insight into how the model identifies column similarities. Additionally, we aim to explore integration with existing data catalog and governance platforms to enable real-time column similarity detection in production big data environments

% reference
\bibliographystyle{IEEEtran}

% \bibliography{reference}
\end{document}